\documentclass[conference]{IEEEtran}
\IEEEoverridecommandlockouts
\usepackage{cite}
\usepackage{amsmath,amssymb,amsfonts}
\usepackage{algorithmic}
\usepackage{graphicx}
\usepackage{textcomp}
\usepackage{xcolor}
\usepackage[numbers]{natbib}
\usepackage{url}
\usepackage{natbib}

\begin{document}

\title{Mitigating Semantic Drift: Evaluating LLMs' Efficacy in Psychotherapy through MI Dialogue Summarization} 


\author{\IEEEauthorblockN{Vivek Kumar}
\IEEEauthorblockA{\textit{Research Institute CODE} \\
\textit{University of the Bundeswehr}\\
Munich, Germany \\
vivek.kumar@unibw.de}
\and
\IEEEauthorblockN{Pushpraj Singh Rajawat}
\IEEEauthorblockA{\textit{Department of Psychology} \\
\textit{Barkatullah University}\\
Bhopal, India \\
psr@manoraksha.com}
\and
\IEEEauthorblockN{Eirini Ntoutsi}
\IEEEauthorblockA{\textit{Research Institute CODE} \\
\textit{University of the Bundeswehr}\\
Munich, Germany \\
eirini.ntoutsi@unibw.de}
}



\maketitle

\begin{abstract}
Recent advancements in large language models (LLMs) have shown their potential across both general and domain-specific tasks. However, there is a growing concern regarding their lack of sensitivity, factual incorrectness in responses, inconsistent expressions of empathy, bias, hallucinations, and overall inability to capture the depth and complexity of human understanding, especially in low-resource and sensitive domains such as psychology. To address these challenges, our study employs a mixed-methods approach to evaluate the efficacy of LLMs in psychotherapy. We use LLMs to generate precise summaries of motivational interviewing (MI) dialogues and design a two-stage annotation scheme based on key components of the Motivational Interviewing Treatment Integrity (MITI) framework, namely evocation, collaboration, autonomy, direction, empathy, and a non-judgmental attitude. Using expert-annotated MI dialogues as ground truth, we formulate multi-class classification tasks to assess model performance under progressive prompting techniques, incorporating one-shot and few-shot prompting. Our results offer insights into LLMs' capacity for understanding complex psychological constructs and highlight best practices to mitigate ``semantic drift" in therapeutic settings. Our work contributes not only to the MI community by providing a high-quality annotated dataset to address data scarcity in low-resource domains but also critical insights for using LLMs for precise contextual interpretation in complex behavioral therapy.
\end{abstract}


\section{Introduction}
Integrating Large Language Models (LLMs) into mental healthcare has garnered significant attention due to their potential to enhance diagnostics, therapeutic interventions, and patient engagement and, to serve  as extra support for clinicians. State-of-the-art (SOTA) LLMs have demonstrated capabilities in understanding and generating human-like text, which can be instrumental in catering to e-health services. However, unlike humans, LLMs lack the innate ability to infer specialized domain knowledge, which makes their deployment in mental health and, in general, in healthcare challenging. Additionally, LLMs often exhibit issues such as inconsistencies in generated text (e..g, contradictory outputs), semantic drift (gradual deviation from the intended topic or context), hallucinations (generation of factually incorrect or fabricated information), and susceptibility to various biases (reflecting or amplifying societal stereotypes present in training data). These shortcomings can lead to outputs that inadvertently reinforce stereotypes or provide unequal care \citep{10.1093/jamia/ocad258,liu-etal-2024-lost}. A further challenge is the scarcity of high-quality, annotated datasets specific to mental health, which hampers the effective training and fine-tuning of LLMs to produce accurate and contextually appropriate responses \cite{Yuan2023-tg}. Therefore, this study investigates the performance of LLMs in complex and low-resource domains such as mental health to address these challenges. We evaluate three LLMs:  DeepSeek\footnote{\url{https://www.deepseek.com/}}, ChatGPT\footnote{\url{https://platform.openai.com/docs/models/overview}} and Gemini\footnote{https://deepmind.google/technologies/gemini/}, across six multi-class classification problem tasks, including precise contextual summary generation and annotation of MI dialogues. Additionally, we develop a summary annotation scheme based on Motivational Interviewing Treatment Integrity (MITI)\footnote{\url{https://casaa.unm.edu/tools/miti.html}} scheme \citep{moyers2014motivational,MOYERS201636} to assess the quality of the generated summaries of MI dialogues across six psychotherapy components. Overall, our contributions are as follows: 
\begin{itemize}
    \item \textbf{Annotation scheme}: We propose an annotation scheme grounded in the MITI coding system to effectively capture the nuances of MI dialogues in psychotherapy. 
    \item \textbf{AnnoSUM-MI}: We introduce the \textbf{AnnoSUM-MI} dataset, expert-annotated across six key MITI dimensions: evocation, collaboration, autonomy, direction, empathy, and non-judgmental attitude. 
    \item \textbf{Heuristic prompting approach}: We propose progressive prompting techniques using one- and few-shot prompt augmentation strategies to guide LLMs in generating contextual MI dialogue summaries.
    \item \textbf{LLMs evaluation}: We evaluate three SOTA LLMs, namely  DeepSeek, ChatGPT, and Gemini, w.r.t. their efficacy in (a) understanding the complexity of MI dialogues, (b) generating summaries of MI dialogues, and (iii) using them as a tool to perform automated data annotation. 
    \item \textbf{Reproducibility}: The AnnoSUM-MI dataset is made publicly available at\footnote{\url{https://github.com/vsrana-ai/AnnoSUM-MI}}.
\end{itemize}

The rest of the paper is organized as follows: Related work is discussed in Section~\ref{lit_survey}. Section~\ref{prob_stat} presents the AnnoMI dataset, the proposed annotation scheme used to generate the AnnoSUM-MI dataset, and the problem statement. Section~\ref{meth} describes the methodology and experimental design. The evaluation results are presented in Section\ref{result}. Section~\ref{conc} concludes our work and outlines directions for future work.

\section{Related Work}\label{lit_survey}
Recent studies have underscored the persistent challenges in applying machine learning (ML), particularly large language models (LLMs), to the mental health domain due to critical data scarcity \citep{gao-etal-2023-enabling}. Also, difficulty in acquiring large-scale, clinically validated datasets due to privacy concerns, cost of employing human-experts and the inherent sensitive nature of mental health discourse, is another significant challenge. The scarcity not only hampers the development of robust ML systems but also compromises the accuracy of LLM evaluation and benchmarking, thereby limiting the generalizability and fairness of employed models \citep{Sogancioglu_Mosteiro_Salah_Scheepers_Kaya_2024,balloccu-etal-2024-leak,balloccu-etal-2024-ask}. To address this gap, researchers have focused on the development and public release of expert-annotated mental health datasets. Notable examples include the Motivational Interviewing (MI) Dataset (MI Dataset)\citep{welivita-pu-2022-curating}, AnnoMI \citep{9746035,fi15030110,10.1007/978-3-031-37249-0_10}, BiMISC \citep{sun-etal-2024-eliciting}, and Prompt-Aware Margin Ranking (PAIR) \citep{min-etal-2022-pair,min-etal-2023-verve} and subsequent studies leveraging these resources \citep{sdaih23,10.1007/978-3-031-37249-0_10,kumar2024unlockingllmsaddressingscarce,kumar-etal-2024-unlocking}. 

While these datasets have facilitated the research in low-resource sensitive domains such as mental health, the continued lack of clinically grounded, representative, and scalable data remains a significant barrier. For example, the MI Dataset \citep{welivita-pu-2022-curating} lacks annotations for client behaviors and each utterance is labeled to a single code, limiting its utility for nuanced analysis. Similarly, the BiMISC dataset \citep{sun-etal-2024-eliciting} relies on inferred rather than clinically verified diagnoses, which may reduce its applicability in real-world clinical settings. In response to these limitations, this study makes a critical contribution to the SOTA and beyond by introducing a multiscale annotated dataset, implementing a rigorous annotation scheme, and proposing robust evaluation frameworks for measuring semantic drift in LLMs.

\section{Dataset, Problem Statement and Annotation Scheme}\label{prob_stat}

We build on the \textbf{AnnoMI}\footnote{\url{https://github.com/vsrana-ai/BIAS-FairAnnoMI}}, a dataset compliant with the General Data Protection Regulation (GDPR) of 131 faithfully transcribed and expert-annotated demonstrations of high- and low-quality MI, an effective therapy strategy that evokes client motivation for positive change. The dialogues describe talk turns between the client and the therapist \citep{9746035,fi15030110,10.1007/978-3-031-37249-0_10} and there are 108 high- and 23 low-quality MI. The data is unique in the context that it demonstrates the optimal behavioral therapy practices by drawing a contrasting distinction between high- and low-quality MI sessions.

\subsection{Annotation Scheme}\label{annotation}
The annotation scheme is developed by experts strictly adhering to the Motivational Interviewing Treatment Integrity (MITI)\footnote{\url{https://casaa.unm.edu/tools/miti.html}} scheme \citep{moyers2014motivational,MOYERS201636}. It consists of six dimensions/components: Evocation, Collaboration, Autonomy, Direction, Empathy and Non-judgmental attitude. Out of six dimensions, the first five dimensions are essentially the components of the MITI framework, and the sixth one, \textbf{Non-Judgmental Attitude}, is included to extend the annotation scope and better reflect the skillset required in MI-based therapeutic sessions. 

These six dimensions were selected meticulously to guide the MI dialogue summary generation process through LLMs in a way that minimizes \emph{semantic drift}, i.e., the tendency of generated text to gradually diverge from the original meaning or intent of the source dialogue, and enhances \emph{contextual fidelity}, i.e., the preservation of the therapeutic tone, intent, and relational dynamics inherent to motivational interviewing. We briefly describe each dimension below:
\begin{itemize}
    \item \textbf{Evocation:} This attribute is central to MI because it focuses on drawing out the client's motivations and reasons for change, a key element in developing intrinsic motivation. Without evocation, MI risks becoming directive rather than maintaining its client-centered therapeutic approach.
    \item \textbf{Collaboration:} MI is built on a relationship between the therapist and client, where both work through change talk. This attribute ensures that the therapist maintains a collaborative, rather than authoritative, stance in the therapy session.
    \item \textbf{Autonomy:} Respecting the client’s independence is crucial in MI. By emphasizing autonomy, the therapist supports the client’s ability to make their own decisions, which is empowering and helps build the client’s confidence in their ability to change.
    \item \textbf{Direction:} While MI is client-centered, the therapist still needs to guide the conversation in a constructive way. This attribute ensures that the therapist maintains focus on the client’s goals without being too rigid or prescriptive.
    \item \textbf{Empathy:} Understanding and connecting with the client's feelings and perspectives is fundamental to building trust and rapport in MI. High levels of empathy contribute to a more supportive and effective therapeutic relationship.
    \item \textbf{Non-Judgmental Attitude:} We added this attribute because it plays a crucial role in creating a safe and accepting environment for the client. MI thrives on the therapist’s ability to listen without passing judgment, allowing the client to explore their thoughts and feelings openly. A non-judgmental attitude helps prevent the client from feeling criticized or defensive, which can hinder progress. This attribute complements the MITI framework by reinforcing the importance of an unbiased and supportive therapeutic environment. 
\end{itemize}

We utilize a five-point Likert scale to assess attitudes, perceptions, and behaviors across the six attributes, which are essentially the guiding principles of the MITI tool:
\begin{itemize}
    \item \textbf{Extremely Low (1):} The attribute is almost entirely absent in the conversation, with minimal or no evidence of its presence.
    \item \textbf{Low (2):} The attribute is present but only weakly demonstrated, with limited influence on the conversation. 
    \item \textbf{Moderate (3):} The attribute is somewhat evident and has a moderate impact on the conversation.
    \item \textbf{High (4):} The attribute is strongly demonstrated and positively influences the conversation.
    \item \textbf{Extremely High (5):} The attribute is strongly demonstrated and is a key driver of the conversation's success.
\end{itemize}

The annotation scheme is applied to both the original MI dialogues 
and the LLM-generated dialogue summaries. 
The process is detailed in Section~\ref{meth}.


\subsection{Problem Statement}
In this work, we use LLMs to generate contextual summaries of MI dialogues and subsequently classify them along the six target dimensions of the proposed annotation scheme: evocation, collaboration, autonomy, direction, empathy, and non-judgmental attitude, using five-point Likert-scale values.
We formulate this as a \emph{multi-output multi-class classification} problem, where each summary is independently evaluated across all six dimensions. For each dimension, the model assigns a discrete class label from the Likert scale \{1, 2, 3, 4, 5\}, indicating the degree to which the corresponding dimension is expressed in the summary. 

To evaluate the capabilities of LLMs in this context, we address the following research questions:\\
\textbf{RQ1}: How effective are LLMs in accurately summarizing complex MI dialogues using guided prompting? \\
\textbf{RQ2}: To what extent do one-shot and few-shot prompting impact contextual understanding of LLMs and mitigate semantic drift? \\
\textbf{RQ3}: Are LLMs sufficiently reliable to be used for automated dataset annotation in sensitive and low-resource domains such as mental health?




\section{Two stage annotation and evaluation framework}\label{meth}
To evaluate the effectiveness of LLMs in interpreting MI dialogues, generating summaries, and serving as a potential tool for automated annotation, we leverage three LLMs: OpenAI ChatGPT (4.0), Google Gemini (2.0 Flash), and DeepSeek (V3). 

The methodology consists of two main steps: i) Human annotation of original MI dialogues, summary generation using LLMs, and subsequent summary annotation; ii) LLM-based classification of the generated summaries across the six MI dimensions defined in the annotation scheme. 

\subsection{Human annotation of MI dialogues and summaries}
First, the 131 MI sessions in the Anno-MI dataset are expert-annotated by using the proposed annotation scheme (c.f., Section~\ref{annotation}) across six MI dimensions: evocation, collaboration, autonomy, direction, empathy, and non-judgmental attitude (\textbf{Annotation stage 1}). To assess the LLMs' capabilities for MI dialogue summarization, a representative test set of 34 sessions (approximately 25\% of Anno-MI) is selected. We then apply one-shot and few-shot tailored prompting strategies to generate summaries for each of the 34 MI sessions. These prompts are iteratively refined based on expert feedback and observations to enhance summary quality and minimize semantic drift. Once the summaries are generated, a second stage of annotation is conducted using the exact six MI dimensions (\textbf{Annotation stage 2}). Since both annotation stages follow identical criteria, the generated summaries can be directly compared to the original dialogues, enabling both quantitative and qualitative evaluation of LLM performance in capturing MI-specific contextual semantics.  


The 131 motivational interviewing (MI) sessions are divided into training (n = 97) and test (n = 34) sets, each stratified to represent AnnoMI topics proportionally. The training set is comprised of 82 high-quality and 15 low-quality sessions, while the test set includes 26 high-quality and 8 low-quality sessions. We selected 15 MI sessions, spanning both quality levels, to serve as a common subset for annotation. The inter-annotation agreement (Cohen’s Kappa) score for these 15 MI sessions is 0.50. According to standard benchmarks \citep{landis1977measurement}, a score between 0.41 and 0.60 is considered  \emph{moderate}. Given the complexity of the annotation task, due to multi-scale parameters and the multi-class classification setting, the score of 0.52 in our study is consistent with published expectations \citep{TANANA201643,hallgren2012computing}.

\subsection{LLM-based classification of summaries}
\begin{figure*}[htbp]
\centerline{\includegraphics[width=6.5in,keepaspectratio]{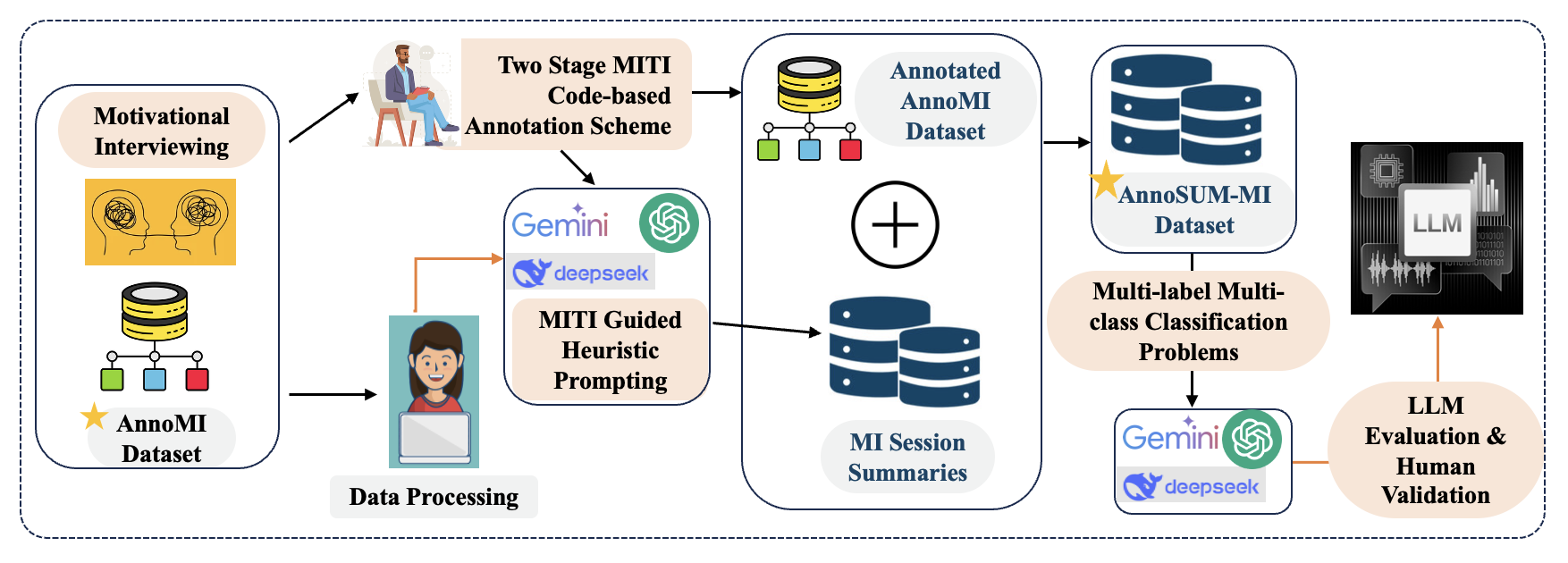}}
\caption{The pipeline showing the annotation protocol and classification using LLMs.}
\label{pipeline}
\end{figure*}

\begin{figure}[htbp]
\centerline{\includegraphics[width=3in,keepaspectratio]{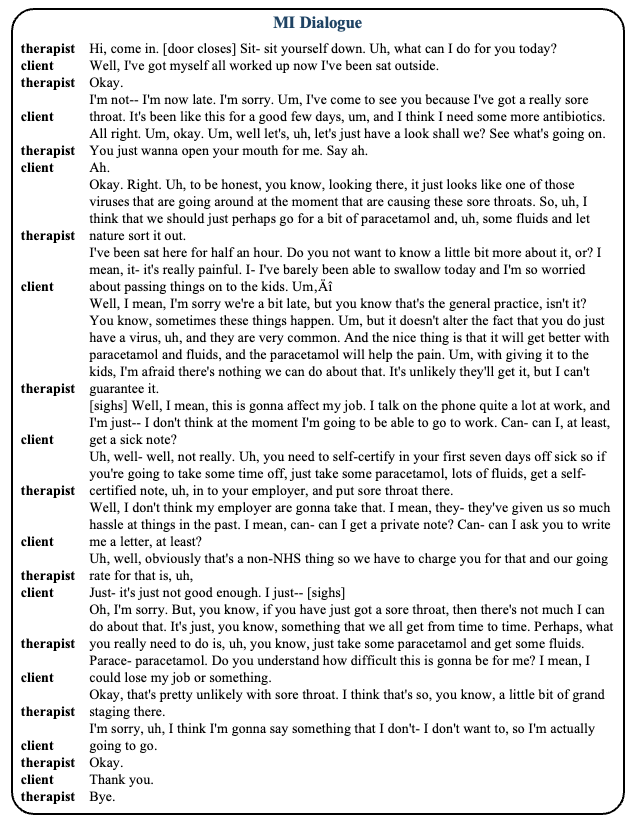}}
\caption{An example MI session.}
\label{original_mi}
\end{figure}

\begin{figure}[htbp]
\centerline{\includegraphics[width=3in,keepaspectratio]{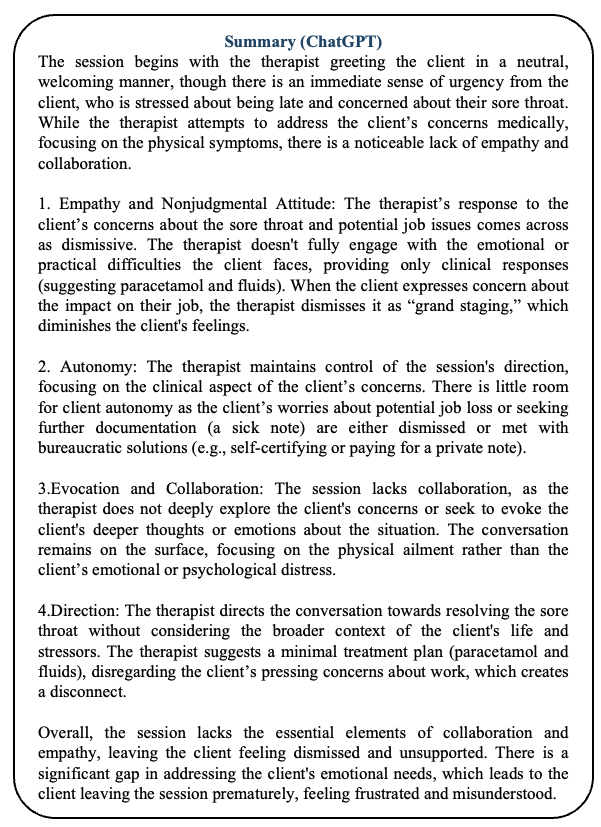}}
\caption{An example MI session summary generated by OpenAI ChatGPT.}
\label{chatgpt_summ}
\end{figure}

\begin{figure}[htbp]
\centerline{\includegraphics[width=3in,keepaspectratio]{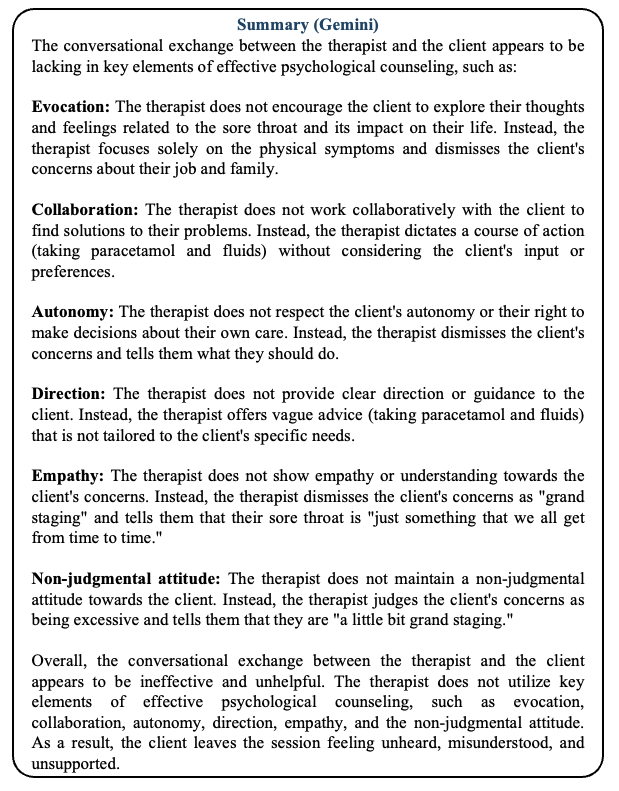}}
\caption{An example MI session summary generated by Google Gemini.}
\label{gemini_summ}
\end{figure}

\begin{figure}[htbp]
\centerline{\includegraphics[width=3in,keepaspectratio]{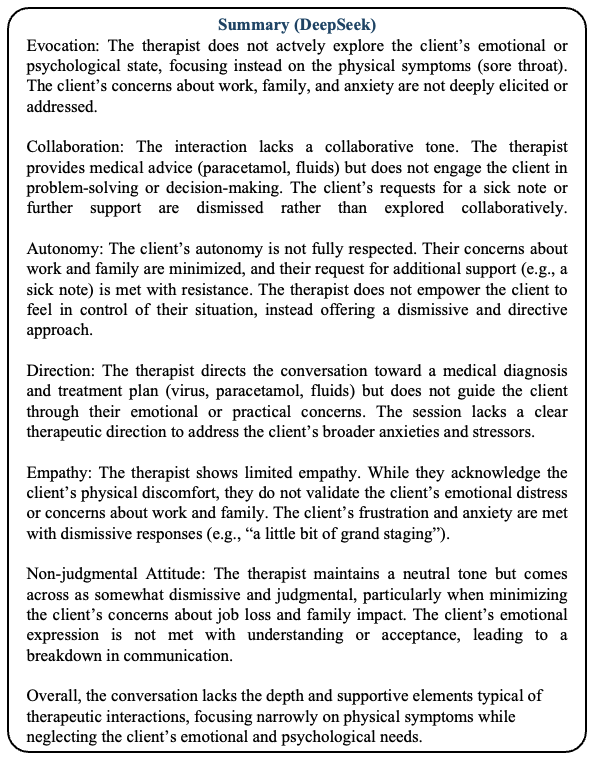}}
\caption{An example MI session summary generated by DeepSeek.}
\label{deepseek_summ}
\end{figure}

\begin{figure}[htbp]
\centerline{\includegraphics[width=3in,keepaspectratio]{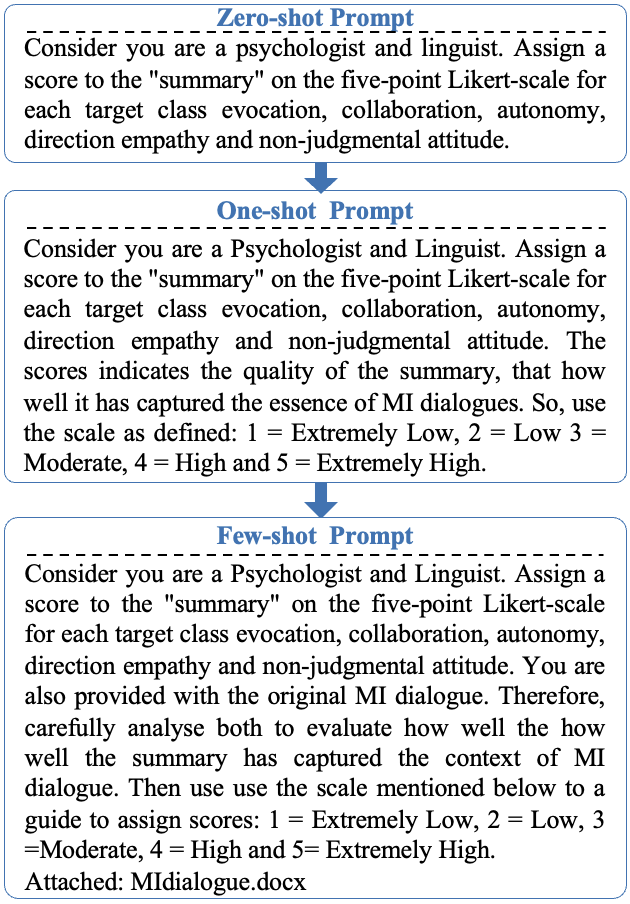}}
\caption{Examples of refined prompts for MI dialogue summarization. The "MIdialogue.docx" here contains the MI sessions for the few-shot prompts.}
\label{prompt}
\end{figure}

In the next step, the LLMs are used to predict Likert-scale scores (1–5) for each of the six MI dimensions: evocation, collaboration, autonomy, direction, empathy, and non-judgmental attitude. An illustration of the process—including an example MI session, the corresponding summaries from ChatGPT, Gemini, and DeepSeek, as well as the different refined prompts, is provided in Fig.~\ref{original_mi}, Fig.~\ref{chatgpt_summ}, Fig.~\ref{gemini_summ}, Fig.~\ref{deepseek_summ}, and Fig.~\ref{prompt}, respectively.  To evaluate LLM performance, we conduct 18 sets of experiments. Specifically, for each of the three LLMs (ChatGPT, Gemini, and DeepSeek), we generate summaries using one-shot and few-shot prompting strategies, resulting in six summaries for each input MI session. Each LLM is then tasked with classifying all six sets covering both its own outputs and those of the other models across the six MI dimensions. For ease of understanding the complete process for annotation and classification is shown in Fig.~\ref{pipeline}.


\section{Evaluation results}\label{result}
\begin{figure}[!htbp]
\centerline{\includegraphics[width=3.5in,keepaspectratio]{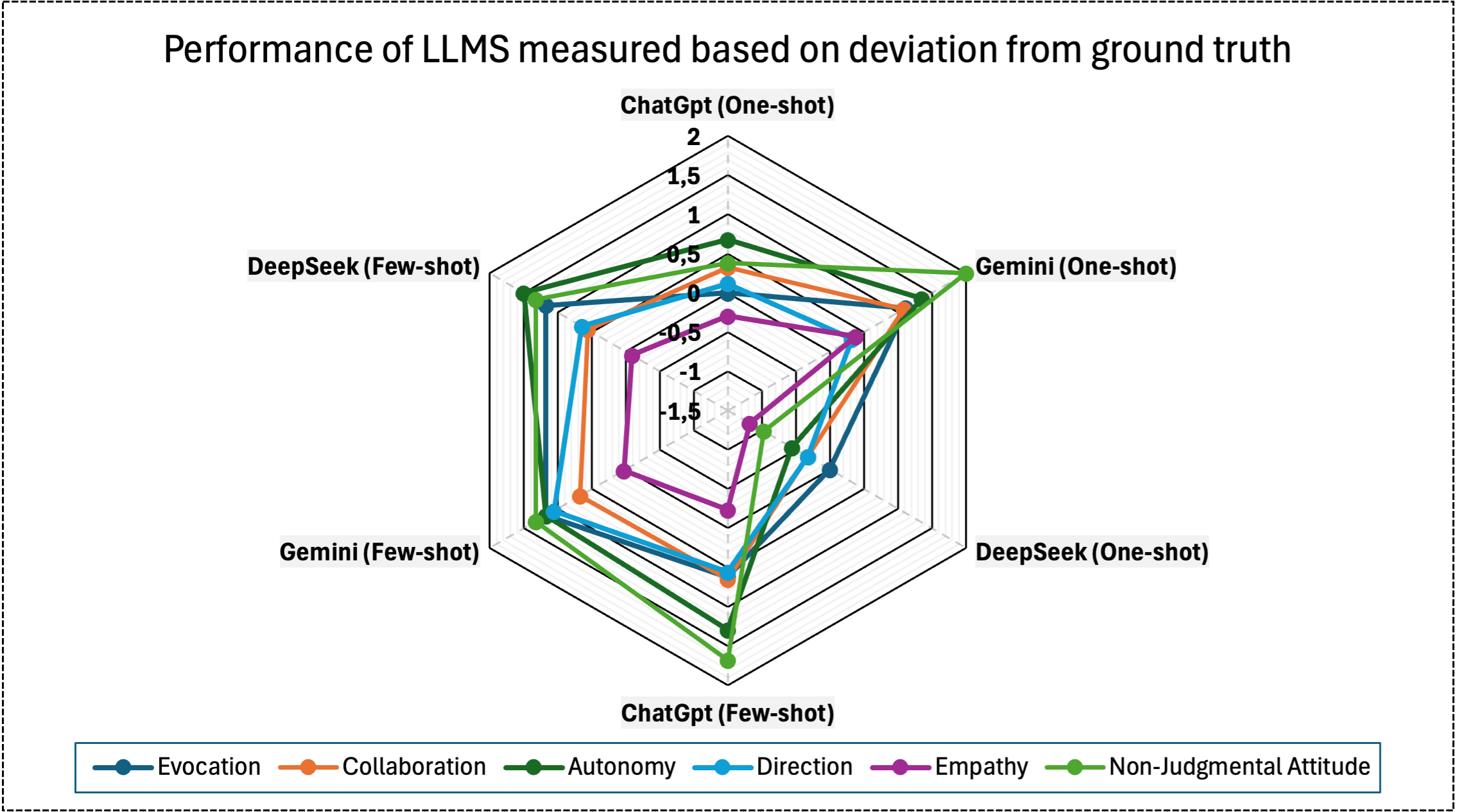}}
\caption{
Radar plot showing the deviation of ChatGPT-based scoring from ground truth across six MI dimensions for summaries generated by ChatGPT, Gemini, and DeepSeek using one- and few-shot prompts. Lower values indicate better alignment with expert annotations.}
\label{radar_shot}
\end{figure}

\begin{figure}[htbp]
\centerline{\includegraphics[width=3.5in,keepaspectratio]{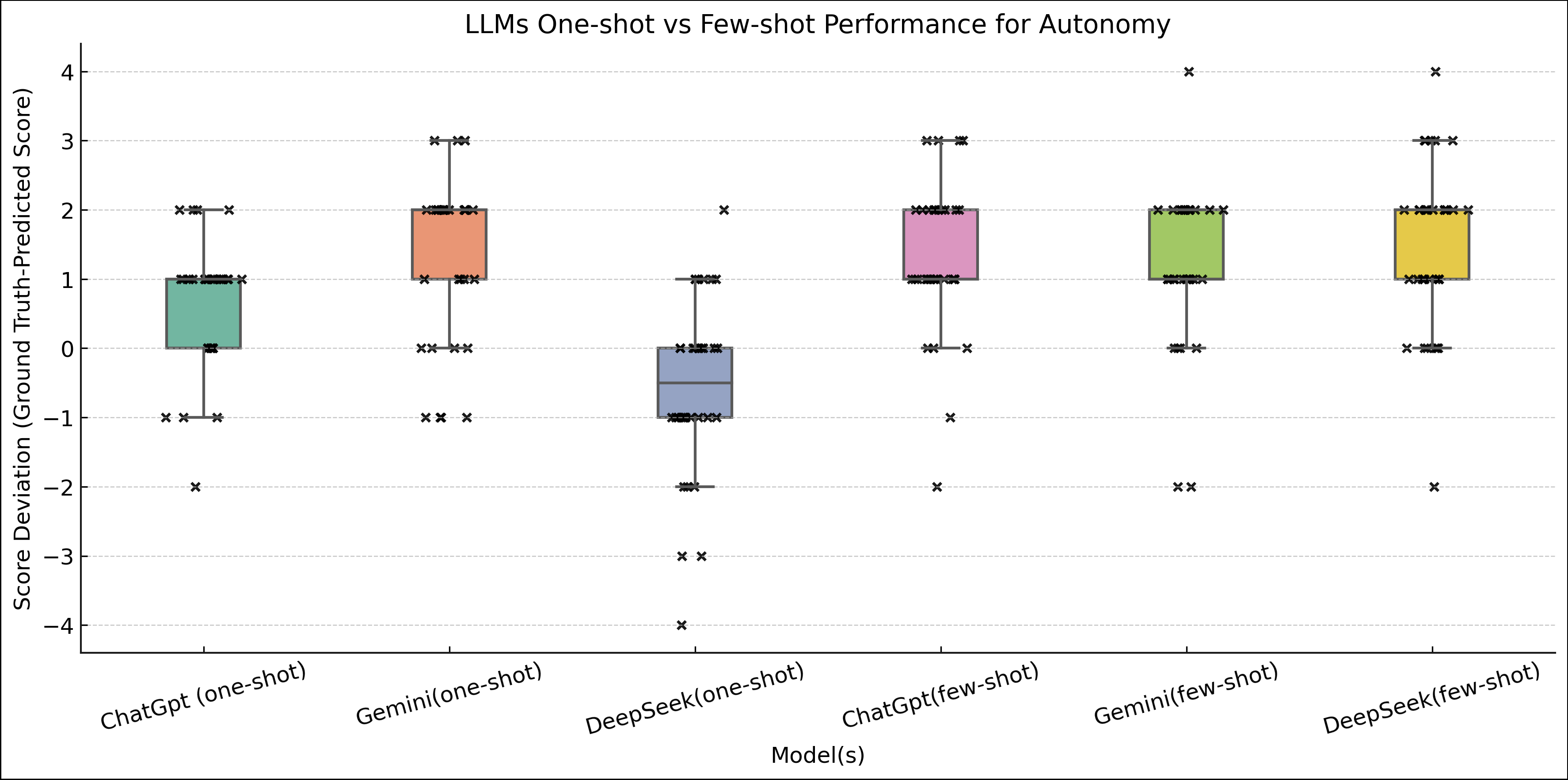}}
\caption{Score density plot for one- and few-shot experiments for attribute Autonomy.}
\label{autonomy}
\end{figure}

\begin{figure}[htbp]
\centerline{\includegraphics[width=3.5in,keepaspectratio]{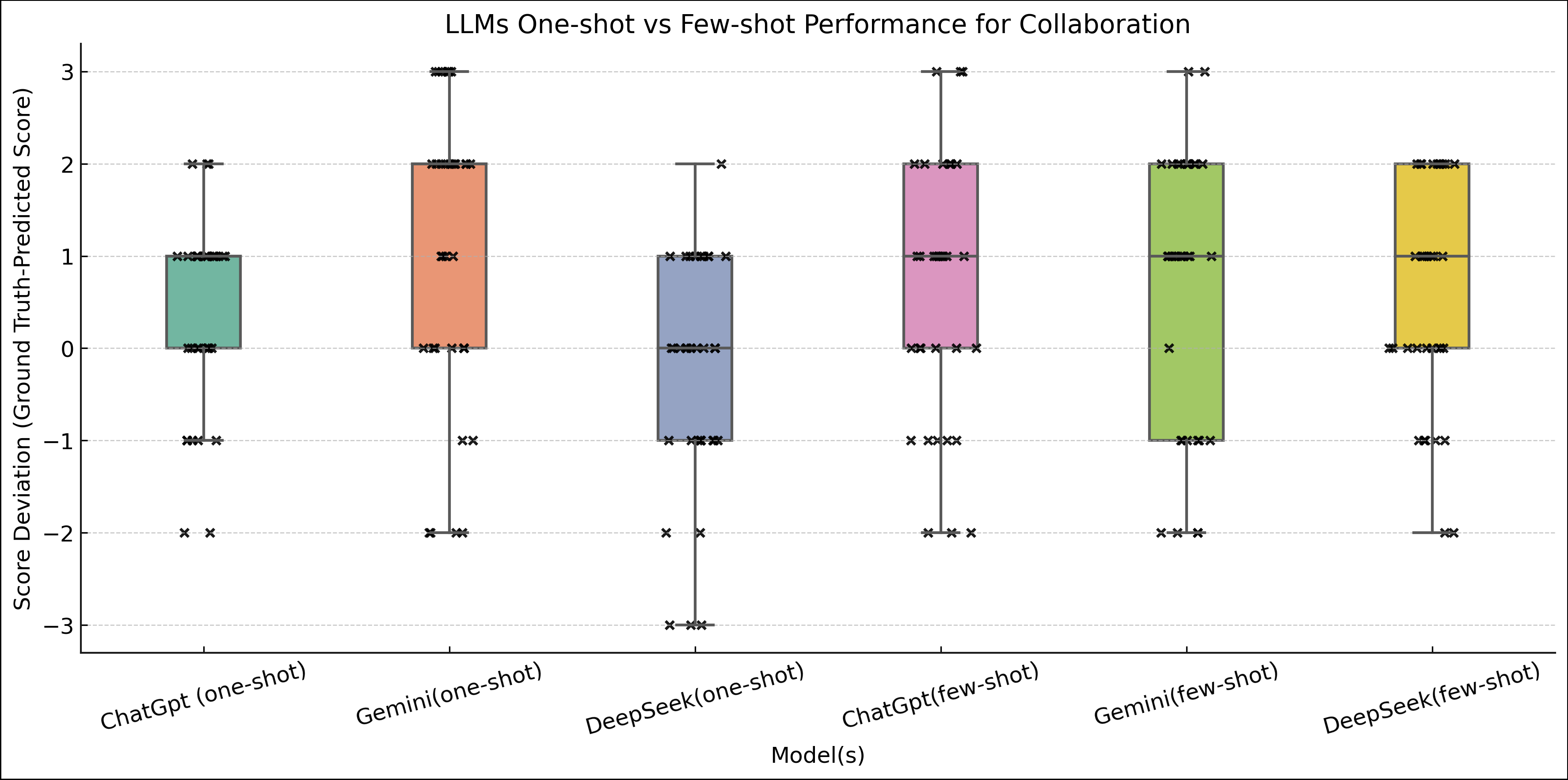}}
\caption{Score density plot for one- and few-shot experiments for attribute Collaboration.}
\label{collaboration}
\end{figure}

\begin{figure}[htbp]
\centerline{\includegraphics[width=3.5in,keepaspectratio]{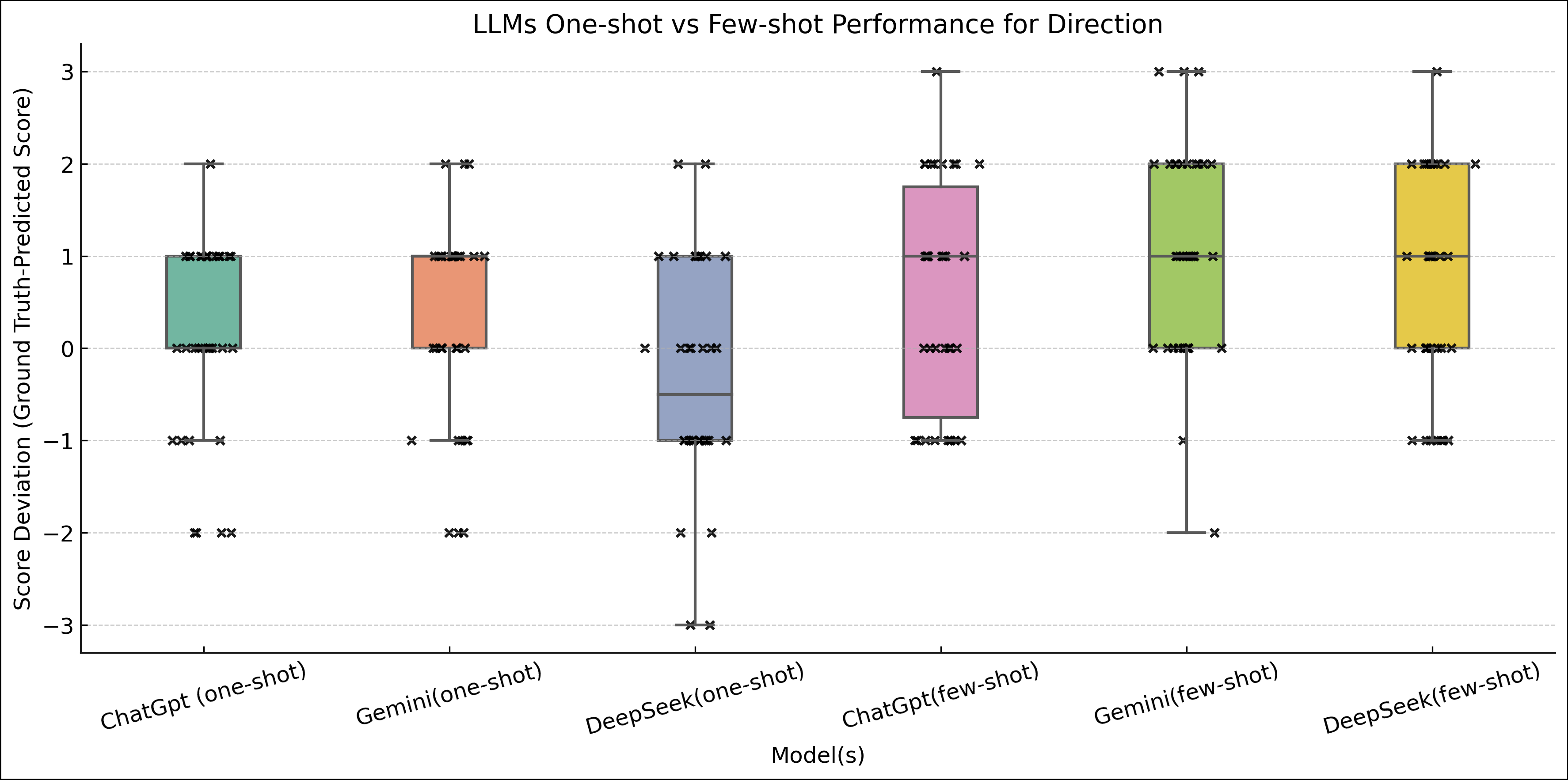}}
\caption{Score density plot for one- and few-shot experiments for attribute Direction.}
\label{direction}
\end{figure}

\begin{figure}[htbp]
\centerline{\includegraphics[width=3.5in,keepaspectratio]{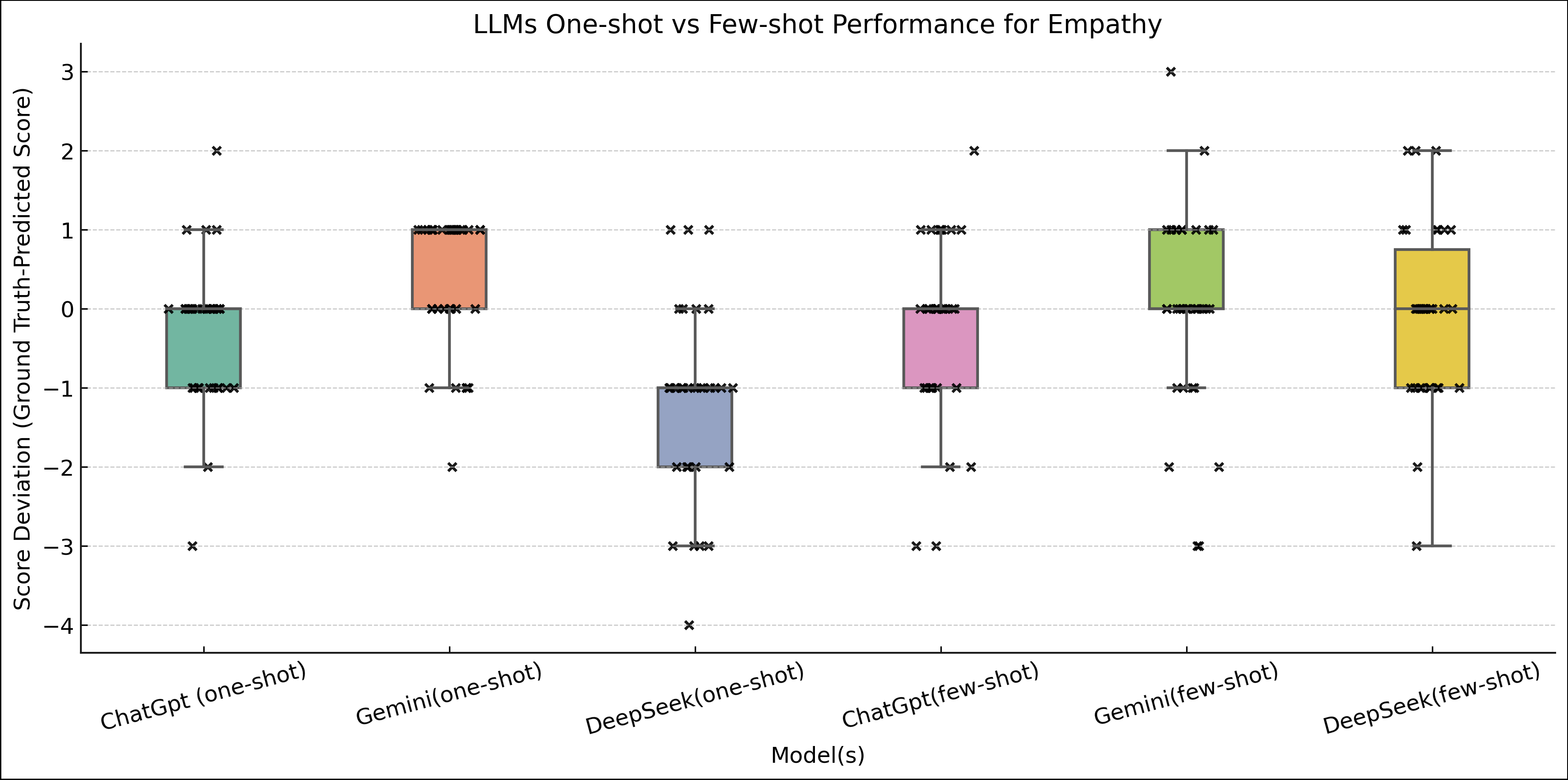}}
\caption{Score density plot for one- and few-shot experiments for attribute Empathy.}
\label{empathy}
\end{figure}

\begin{figure}[htbp]
\centerline{\includegraphics[width=3.5in,keepaspectratio]{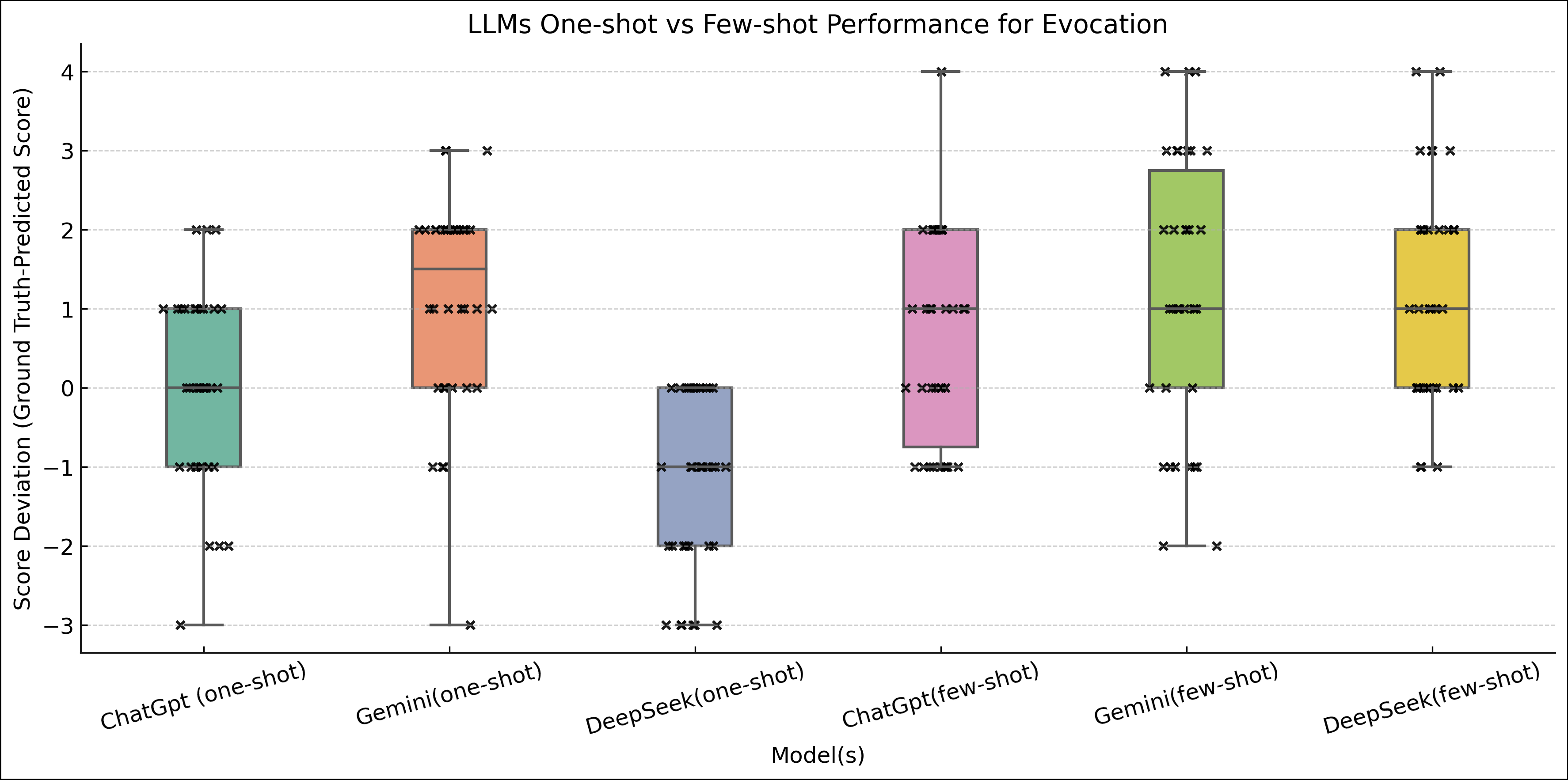}}
\caption{Score density plot for one- and few-shot experiments for attribute Evocation.}
\label{evocation}
\end{figure}

\begin{figure}[htbp]
\centerline{\includegraphics[width=3.5in,keepaspectratio]{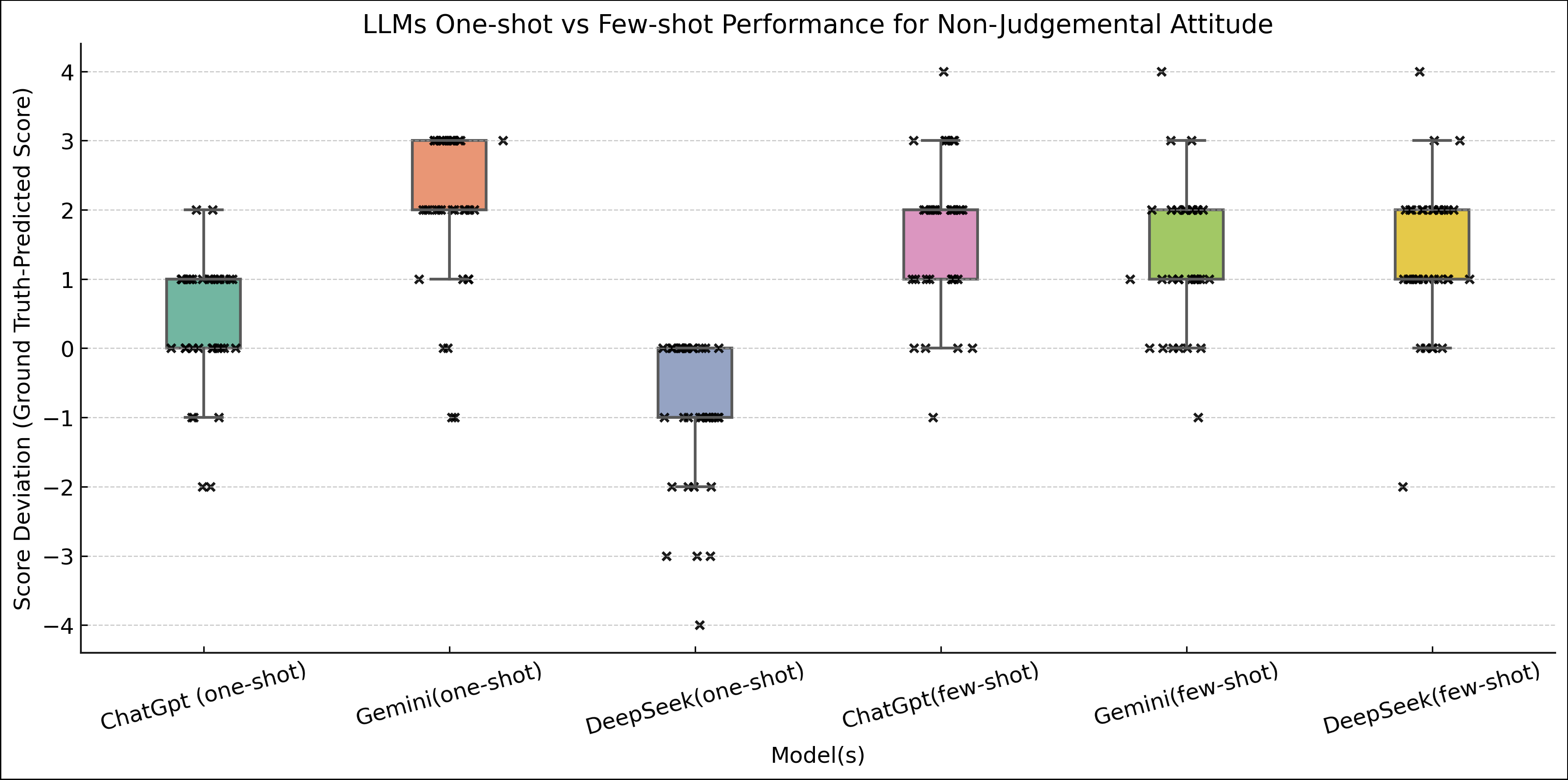}}
\caption{Score density plot for one- and few-shot experiments for attribute Non-Judgmental Attitude.}
\label{non_judgmental}
\end{figure}

Given that the task undertaken in this study is quite challenging and subjective due to the inherent complexity of the target domain, the usual metrics, such as accuracy, precision, recall, and F1-score, are not best suited to measure the performance of LLMs. Instead, we focus on semantic drift and evaluate the deviation of LLM summaries from expert-annotated ground truth. 
The deviation is calculated as the difference between the predicted and ground truth Likert-scale values (ranging from 1 to 5) for each of the six MI dimensions. In Figure~\ref{radar_shot}, we show the deviation of classification performance (using ChatGPT for scoring) 
from ground truth across the six MI dimensions. Results are shown for summaries generated by ChatGPT, Gemini, and DeepSeek using one- and few-shot prompts. The detailed performance for this particular experiment using ChatGPT scoring for one- and few-shot prompts and each attribute is shown in Fig.~\ref{autonomy}, Fig.~\ref{collaboration}, Fig.~\ref{direction}, Fig.~\ref{empathy}, Fig.~\ref{empathy}, Fig.~\ref{evocation}, and Fig.~\ref{non_judgmental}, respectively. The remaining results are available in the official repository\footnote{\url{https://github.com/vsrana-ai/AnnoSUM-MI}}. Due to space constraints, we summarize the main findings below:

\begin{itemize}
\item \textbf{Gemini}: Among all the LLMs used in this study, it has produced the lowest quality summaries for MI dialogues due to the extreme approach to interpreting the components of MI dialogues. That means Gemini's summaries are brief and not detailed enough to reflect the intensity of the emotion involved. This observation is also supported by the experimental outcome of the multi-class classification task, where for both one- and few-shot prompt strategies, Gemini has taken the maximum deviation from the ground truth.  

\item \textbf{DeepSeek}: The latest release, DeepSeek, has performed better than Gemini, but it also has several shortcomings and hallucination problems. For long prompts, it loses context in between; thus, the inconsistencies are visibly notable in the summary outcome. Like Gemini, it also takes an extreme stance to describe the presence of attributes in MI dislodges; however, it is slightly better, as the summaries are still a bit descriptive. 

\item \textbf{ChatGPT}: ChatGPT has shined among all the LLMs and has consistently performed better for all sets of experiments. The best performance of ChatGPT is observed to be with the summaries generated by ChatGPT, where the deviation from ground truth is the least, and the misclassification scores are within the range of +1 and -1. Expert input also asserts that the quality of ChatGPT-generated summaries is reasonable. ChatGPT demonstrated a less extreme approach, adequately descriptive, and less prone to missing out on empathetic notions in MI dialogues. 
\end{itemize}




\section{Conclusion, Limitations and Future Work}\label{conc}
\textbf{Summary of contributions:} In this work \citep{11228771}, we leverage three SOTA LLMs for MI dialogue summarization and explore their potential for automating the annotation of complex MI sessions along six key dimensions inspired by the MITI coding framework. The motivation is to reduce reliance on human experts and accelerate the generation of human-like annotated data, addressing the challenges of data scarcity in low-resource, complex domains. To achieve this, we develop a two-stage comprehensive annotation scheme and evaluate the efficacy of LLMs by formulating the problem as a multi-output, multi-class classification problem. Our results show that ChatGPT (one-shot) demonstrates the closest alignment with the ground truth, while Gemini (one-shot) exhibits the highest overall deviation. Our work also serves as a guideline to develop a robust evaluation scheme grounded on MITI/MISC to establish an annotation protocol for complex utterance and dialogue-level conversational therapy sessions and evaluate the efficacy of LLMs for such talks. \\
\textbf{Limitations \& Future Works:}  In our approach, LLMs are used for both generating and evaluating summaries, which may introduce bias and affect the objectivity of performance assessment. Future work will explore independent human evaluations or cross-model validation to mitigate this. Additionally, we aim to expand our evaluation to include more SOTA LLMs, varied prompting strategies, and additional datasets. We also plan to explore alternative methods to address the problem of data scarcity in low-resource domains.




\section*{Acknowledgment}
This research work is funded by the European Union Horizon Europe Project STELAR, Grant Agreement ID: 101070122.

\bibliographystyle{IEEEtranN} 
\bibliography{biblio}


\end{document}